\documentclass{article}

\usepackage{arxiv}

\usepackage{lmodern}

\usepackage[utf8]{inputenc}
\usepackage[T1]{fontenc}
\usepackage[numbers,sort&compress]{natbib}   
\usepackage{hyperref}
\usepackage{url}
\usepackage{booktabs}
\usepackage{amsmath,amssymb,amsfonts}
\usepackage{nicefrac}
\usepackage{microtype}
\usepackage{graphicx}
\usepackage{doi}
\usepackage{multirow}
\usepackage{xcolor}
\usepackage{bm}
\usepackage{tabularx}
\usepackage{tikz}
\usetikzlibrary{positioning, arrows.meta, shapes.geometric, fit, calc, decorations.pathreplacing}

\makeatletter
\renewcommand{\@maketitle}{%
  \vbox{%
    \hsize\textwidth
    \linewidth\hsize
    \vskip 0.1in
    \@toptitlebar
    \centering
    {\Large\sc \@title\par}%
    \@bottomtitlebar
    \textsc{\undertitle}\\
    \vskip 0.1in
    \def\And{%
      \end{tabular}\hfil\linebreak[0]\hfil%
      \begin{tabular}[t]{c}\bf\rule{\z@}{24\p@}\ignorespaces%
    }%
    \def\AND{%
      \end{tabular}\hfil\linebreak[4]\hfil%
      \begin{tabular}[t]{c}\bf\rule{\z@}{24\p@}\ignorespaces%
    }%
    \begin{tabular}[t]{c}\bf\rule{\z@}{24\p@}\@author\end{tabular}%
    \vskip 0.4in \@minus 0.1in \center{\@date}\vskip 0.2in
  }%
}
\makeatother

\title{SA-CycleGAN-2.5D: Self-Attention CycleGAN with Tri-Planar Context for Multi-Site MRI Harmonization}

\author{
  Ishrith Gowda \\
  Department of Electrical Engineering and Computer Sciences \\
  University of California, Berkeley \\
  Berkeley, CA 94720 \\
  \texttt{ishrithgowda@berkeley.edu}
  \And
  Chunwei Liu \\
  Department of Computer Science \\
  Purdue University \\
  West Lafayette, IN 47907 \\
  \texttt{chunwei@purdue.edu}
}

\date{March 2026}

\renewcommand{\shorttitle}{SA-CycleGAN-2.5D for Multi-Site MRI Harmonization}
\renewcommand{\headeright}{Preprint. Under review at MICCAI 2026}
\renewcommand{\undertitle}{Preprint. Under review at MICCAI 2026}

\hypersetup{
  pdftitle={SA-CycleGAN-2.5D: Self-Attention CycleGAN with Tri-Planar Context
            for Multi-Site MRI Harmonization},
  pdfsubject={cs.CV, eess.IV, cs.LG},
  pdfauthor={Ishrith Gowda, Chunwei Liu},
  pdfkeywords={MRI harmonization, scanner harmonization, multi-site MRI, domain adaptation,
               CycleGAN, generative adversarial network, self-attention, 2.5D, tri-planar,
               glioma, brain tumor, radiomics, BraTS, UPenn-GBM, domain shift,
               maximum mean discrepancy, unpaired image translation},
}

\begin{document}
\maketitle

\begin{abstract}
Multi-site neuroimaging analysis is fundamentally confounded by scanner-induced covariate
shifts, where the marginal distribution of voxel intensities $P(\mathbf{x})$ varies
non-linearly across acquisition protocols while the conditional anatomy
$P(\mathbf{y}|\mathbf{x})$ remains constant. This is particularly detrimental to radiomic
reproducibility, where acquisition variance often exceeds biological pathology variance.
Existing statistical harmonization methods (e.g., ComBat) operate in feature space,
precluding spatial downstream tasks, while standard deep learning approaches are
theoretically bounded by local effective receptive fields (ERF), failing to model the
global intensity correlations characteristic of field-strength bias.

We propose \textbf{SA-CycleGAN-2.5D}, a domain adaptation framework motivated by the
$\mathcal{H}\Delta\mathcal{H}$-divergence bound of Ben-David et al., integrating three
architectural innovations: (1)~A 2.5D tri-planar manifold injection preserving
through-plane gradients $\nabla_z$ at $\mathcal{O}(HW)$ complexity; (2)~A U-ResNet
generator with dense voxel-to-voxel self-attention, surpassing the $\mathcal{O}(\sqrt{L})$
receptive field limit of CNNs to model global scanner field biases; and (3)~A
spectrally-normalized discriminator constraining the Lipschitz constant ($K_D \le 1$) for
stable adversarial optimization. Evaluated on 654 glioma patients across two institutional
domains (BraTS and UPenn-GBM), our method reduces Maximum Mean Discrepancy (MMD) by
$99.1\%$ ($1.729 \to 0.015$) and degrades domain classifier accuracy to near-chance
($59.7\%$). Ablation confirms that global attention is statistically essential
(Cohen's $d{=}1.32$, $p{<}0.001$) for the harder heterogeneous-to-homogeneous translation
direction. By bridging 2D efficiency and 3D consistency, our framework yields voxel-level
harmonized images that preserve tumor pathophysiology, enabling reproducible multi-center
radiomic analysis.
\end{abstract}


\section{Introduction}
\label{sec:intro}

Neurological oncology studies increasingly require aggregating data across institutions
to achieve sufficient statistical power for treatment response modeling, genomic
correlation, and survival analysis. Yet multi-site MRI acquisition introduces systematic
covariate shifts (differences in field strength (1.5T vs.\ 3T), vendor-specific
gradient calibration, and site-specific protocol choices) that confound downstream
analyses. Scanner-induced effects routinely exceed inter-subject biological variability in
glioma cohorts~\cite{fortin2018harmonization}, biasing radiomic signatures, degrading
segmentation models trained across sites, and inflating false-discovery rates in
statistical studies.

Voxel-level image harmonization, producing an image indistinguishable from those
acquired at a reference site while preserving the subject's underlying anatomy and
pathology, offers a principled solution. Unlike feature-space corrections, voxel-level
outputs are directly compatible with all downstream spatial tasks (segmentation,
volumetric analysis, radiomics). The challenge is that site labels are often unavailable
in federated or retrospective settings, paired traveling-subject data is impractical at
scale, and the domain shift involves both global bias-field effects and local contrast
variations, requiring a model with both global receptive fields and structural awareness.

We present \textbf{SA-CycleGAN-2.5D}, addressing these challenges through principled
integration of domain adaptation theory with three targeted architectural innovations.
Our framework motivates adversarial training via the $\mathcal{H}\Delta\mathcal{H}$
divergence bound~\cite{ben2010theory}, incorporates inter-slice context through 2.5D
tri-planar encoding, and overcomes the receptive field limitation of convolutional networks
through dense self-attention mechanisms. To our knowledge, this is the first work to
quantify the statistical contribution of self-attention to MRI harmonization quality using
large-effect-size ablation ($d{=}1.13\text{--}1.32$) across all modalities.

\paragraph{Contributions.}
\begin{enumerate}
  \item \textbf{2.5D tri-planar input}: A SliceEncoder25D concatenating adjacent slices
        across four modalities ($12$-channel)~\cite{roth2014twopointfived},
        preserving inter-slice gradients $\nabla_z$ at 2D cost.
  \item \textbf{U-ResNet with pervasive self-attention}: Self-attention~\cite{zhang2019sagan}
        at three of nine bottleneck blocks plus globally, with 11 CBAM~\cite{woo2018cbam}
        modules throughout, at only 3.4\% parameter overhead (1.2M/35.1M).
  \item \textbf{Multi-axis evaluation}: Cycle consistency, domain separation (classifier +
        MMD + KS), and 512-feature radiomics concordance on 654 subjects across two glioma
        cohorts.
  \item \textbf{Domain adaptation framing}: Adversarial harmonization motivated by the
        $\mathcal{H}\Delta\mathcal{H}$-divergence bound~\cite{ben2010theory}, with
        statistically rigorous ablation using bootstrapped confidence intervals, Bonferroni
        correction, and Cohen's $d$ effect sizes.
\end{enumerate}

\section{Related Work}
\label{sec:related}

\subsection{Statistical Harmonization}

ComBat~\cite{johnson2007combat} models site effects as additive and multiplicative terms
in a linear mixed model, removing them via empirical Bayes estimation. While effective
for batch-effect correction in transcriptomics and later neuroimaging~\cite{fortin2017harmonization},
ComBat fundamentally operates in feature space and cannot produce harmonized images.
Fortin et al.~\cite{fortin2017harmonization} adapted ComBat for diffusion MRI metrics and
cortical thickness, and CovBat~\cite{chen2022covbat} extended it to covariance
harmonization. All statistical methods share
two critical limitations: they require explicit site labels (unavailable in federated
learning), and they cannot reconstruct spatially harmonized images required by
segmentation-based downstream tasks.

\subsection{Deep Learning Harmonization}

Supervised approaches such as DeepHarmony~\cite{dewey2019deepharmony} leverage paired
traveling-subject scans to directly learn a voxel-wise correction field, but the logistic
overhead of multi-site traveling-subject acquisition limits scalability. Unpaired methods
based on CycleGAN~\cite{goodfellow2014gan,zhu2017cyclegan} circumvent this requirement; Modanwal et
al.~\cite{modanwal2020mri} and Zhao et al.~\cite{zhao2019harmonization} demonstrated the
viability of cycle-consistent translation for MRI harmonization, but these methods rely
on purely convolutional generators whose effective receptive fields scale as
$\mathcal{O}(\sqrt{L})$~\cite{luo2016erf}, which is insufficient to model global field-strength
biases. Contrastive unpaired translation (CUT)~\cite{park2020cut} improves patch-level
fidelity but similarly lacks global context.

Information-theoretic disentanglement approaches, CALAMITI~\cite{zuo2021calamiti} and
HACA3~\cite{zuo2023haca3}, separate anatomy from contrast encoding using variational
bounds, achieving strong performance but requiring multi-contrast inputs and complex
encoder-decoder pipelines that limit clinical deployment. Adversarial unlearning
approaches~\cite{dinsdale2021deep} produce site-invariant features but no harmonized
images, making them incompatible with spatial downstream tasks.

\subsection{Self-Attention and Transformers in Medical Imaging}

The limitations of local receptive fields in convolutional networks have driven adoption
of attention mechanisms for medical image analysis. SAGAN~\cite{zhang2019sagan}
introduced self-attention into generative networks, enabling long-range feature
interactions. TransUNet~\cite{chen2021transunet} augmented U-Net skip connections with
transformer encoders for segmentation, while Swin-UNet~\cite{cao2022swinunet} replaced
convolutions entirely with shifted-window self-attention, achieving state-of-the-art
results on medical segmentation benchmarks. However, pure transformers face quadratic
attention complexity, making them computationally prohibitive for high-resolution
volumetric processing. Convolutional block attention modules (CBAM)~\cite{woo2018cbam}
provide a lightweight alternative with channel-spatial recalibration at negligible
overhead.

SA-CycleGAN-2.5D occupies a complementary position: it retains the computational
efficiency of convolutions for the bulk of processing, inserting full self-attention only
at the bottleneck where spatial resolution is lowest, while CBAM modules provide
lightweight recalibration at every stage. This hybrid design achieves global context at
manageable computational cost, which we quantify in Section~\ref{sec:experiments}.

\subsection{Domain Adaptation Theory}

The $\mathcal{H}\Delta\mathcal{H}$-divergence bound of Ben-David et al.~\cite{ben2010theory}
provides a theoretical foundation connecting domain shift to target task performance,
motivating the reduction of domain divergence as a principled objective. This bound has
been applied to medical imaging domain adaptation for classification~\cite{dinsdale2021deep},
but its application as a motivating framework for voxel-level image harmonization,
where the generator serves as an empirical divergence-minimizing transport map, has not
been previously formalized.

\section{Methods}
\label{sec:methods}

\subsection{Problem Formulation}

Let $\mathcal{D}_A = \{x_i^A\}_{i=1}^{N_A}$ and $\mathcal{D}_B = \{x_j^B\}_{j=1}^{N_B}$
be unpaired samples from source domain $A$ (multi-site BraTS) and target domain $B$
(single-site UPenn-GBM), respectively. Each sample $x \in \mathbb{R}^{4 \times H \times W}$
is a four-modality (T1, T1CE, T2, FLAIR) 2D slice. We seek generators
$G_{A \to B}: \mathcal{X}_A \to \mathcal{X}_B$ and $G_{B \to A}: \mathcal{X}_B \to \mathcal{X}_A$
that satisfy three criteria:
\begin{enumerate}
  \item \textbf{Domain alignment:} $G_{A\to B}(x^A)$ is indistinguishable from samples
        of $\mathcal{D}_B$ (reduce domain divergence).
  \item \textbf{Cycle consistency:} $G_{B\to A}(G_{A\to B}(x^A)) \approx x^A$ (preserve
        content).
  \item \textbf{Anatomical fidelity:} Tumor morphology, gray-white matter contrast, and
        ventricular structure are preserved under transport.
\end{enumerate}
Critically, no paired data, site labels, or traveling subjects are required.

\subsection{Theoretical Motivation}
\label{sec:theory}

We frame harmonization through the domain adaptation bound of Ben-David et
al.~\cite{ben2010theory}: for any hypothesis $h \in \mathcal{H}$, the target risk is
bounded by:
\begin{equation}
  \epsilon_\mathcal{T}(h) \leq \epsilon_\mathcal{S}(h)
    + \tfrac{1}{2}\, d_{\mathcal{H}\Delta\mathcal{H}}(\mathcal{D}_\mathcal{S}, \mathcal{D}_\mathcal{T})
    + \lambda^*,
  \label{eq:da_bound}
\end{equation}
where $d_{\mathcal{H}\Delta\mathcal{H}}$ is the $\mathcal{H}\Delta\mathcal{H}$-divergence
measuring distributional discrepancy under the hypothesis class, and $\lambda^*$ is the
error of the optimal joint hypothesis. This bound implies that reducing the divergence
between the source and target distributions in voxel space will improve downstream task
generalization on the target domain.

Our adversarial discriminator $D_B$ is trained to distinguish real target-domain samples
from generated ones. By simultaneously training $G_{A\to B}$ to fool $D_B$, we minimize
an adversarial loss that serves as an empirical proxy for domain divergence. The generator
does not directly optimize the $\mathcal{H}\Delta\mathcal{H}$-divergence (a theoretically
intractable quantity), but the adversarial game provides an implicit gradient signal
toward distributions with lower divergence, without requiring paired data or site labels.

Convolutional generators are further bounded by their effective receptive field
(ERF)~\cite{luo2016erf}: the ERF grows as $\mathcal{O}(\sqrt{L})$ with network depth $L$,
meaning that at practical depths the generator cannot correct global field-strength biases
affecting the entire field-of-view in a single forward pass. Self-attention
resolves this by computing pairwise affinities across all $N = H \times W$ spatial
positions, providing direct gradient paths $\partial\mathcal{L}/\partial \mathbf{h}_j$
between any positions $(i,j)$ regardless of spatial distance.

\subsection{Architecture}
\label{sec:arch}

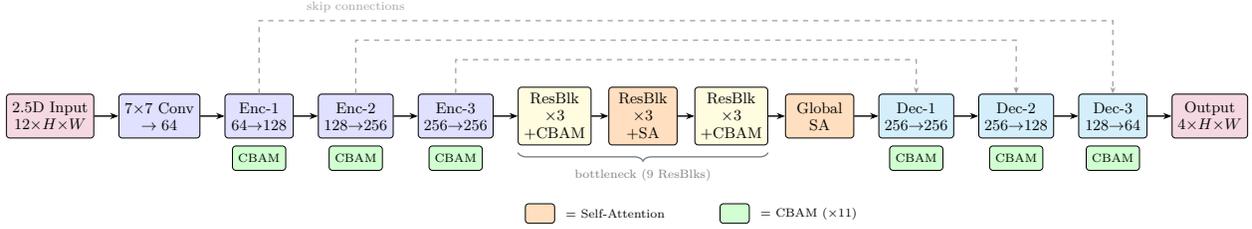
\begin{figure}[t]
  \centering
  \resizebox{\textwidth}{!}{%
  \begin{tikzpicture}[
    block/.style={rectangle, draw, fill=#1, minimum height=0.9cm, minimum width=1.4cm,
                  font=\footnotesize, align=center, rounded corners=2pt},
    block/.default={blue!12},
    sablock/.style={rectangle, draw, fill=orange!25, minimum height=0.9cm, minimum width=1.4cm,
                    font=\footnotesize, align=center, rounded corners=2pt},
    cbam/.style={rectangle, draw, fill=green!18, minimum height=0.5cm, minimum width=0.9cm,
                 font=\scriptsize, align=center, rounded corners=2pt},
    arr/.style={-{Stealth[length=5pt]}, thick},
    skip/.style={-{Stealth[length=4pt]}, thick, dashed, gray!70},
    label/.style={font=\scriptsize, text=gray!70},
  ]
    \node[block=purple!15] (input) {2.5D Input\\$12{\times}H{\times}W$};

    \node[block, right=0.5cm of input] (stem) {$7{\times}7$ Conv\\$\to 64$};

    \node[block, right=0.5cm of stem] (e1) {Enc-1\\$64{\to}128$};
    \node[cbam, below=0.15cm of e1] (c1) {CBAM};
    \node[block, right=0.5cm of e1] (e2) {Enc-2\\$128{\to}256$};
    \node[cbam, below=0.15cm of e2] (c2) {CBAM};
    \node[block, right=0.5cm of e2] (e3) {Enc-3\\$256{\to}256$};
    \node[cbam, below=0.15cm of e3] (c3) {CBAM};

    \node[block=yellow!15, right=0.5cm of e3] (bg1) {ResBlk\\$\times$3\\+CBAM};
    \node[sablock, right=0.35cm of bg1] (bg2) {ResBlk\\$\times$3\\+SA};
    \node[block=yellow!15, right=0.35cm of bg2] (bg3) {ResBlk\\$\times$3\\+CBAM};

    \node[sablock, right=0.35cm of bg3] (gsa) {Global\\SA};

    \node[block=cyan!15, right=0.5cm of gsa] (d1) {Dec-1\\$256{\to}256$};
    \node[cbam, below=0.15cm of d1] (c4) {CBAM};
    \node[block=cyan!15, right=0.5cm of d1] (d2) {Dec-2\\$256{\to}128$};
    \node[cbam, below=0.15cm of d2] (c5) {CBAM};
    \node[block=cyan!15, right=0.5cm of d2] (d3) {Dec-3\\$128{\to}64$};
    \node[cbam, below=0.15cm of d3] (c6) {CBAM};

    \node[block=purple!15, right=0.5cm of d3] (output) {Output\\$4{\times}H{\times}W$};

    \draw[arr] (input) -- (stem);
    \draw[arr] (stem) -- (e1);
    \draw[arr] (e1) -- (e2);
    \draw[arr] (e2) -- (e3);
    \draw[arr] (e3) -- (bg1);
    \draw[arr] (bg1) -- (bg2);
    \draw[arr] (bg2) -- (bg3);
    \draw[arr] (bg3) -- (gsa);
    \draw[arr] (gsa) -- (d1);
    \draw[arr] (d1) -- (d2);
    \draw[arr] (d2) -- (d3);
    \draw[arr] (d3) -- (output);

    \draw[skip] (e3.north) -- ++(0,0.7) -| (d1.north);
    \draw[skip] (e2.north) -- ++(0,1.1) -| (d2.north);
    \draw[skip] (e1.north) -- ++(0,1.5) -| (d3.north);

    \node[label, above=1.55cm of e2] {skip connections};

    \draw[decorate, decoration={brace, amplitude=5pt, mirror}, thick, gray]
      (bg1.south west) ++(0,-0.15) -- ($(bg3.south east)+(0,-0.15)$)
      node[midway, below=0.2cm, font=\scriptsize, text=gray] {bottleneck (9 ResBlks)};

    \node[sablock, minimum height=0.4cm, minimum width=0.6cm] at ($(bg1.south)+(-0.3,-1.4)$) (leg1) {};
    \node[font=\scriptsize, right=0.1cm of leg1] (leg1txt) {= Self-Attention};
    \node[cbam, minimum height=0.4cm, minimum width=0.6cm, right=1.0cm of leg1txt] (leg2) {};
    \node[font=\scriptsize, right=0.1cm of leg2] {= CBAM ($\times$11)};

  \end{tikzpicture}%
  }
  \caption{SA-CycleGAN-2.5D generator architecture. The 2.5D tri-planar input
    ($12$ channels) passes through a convolutional stem, three encoder stages with CBAM
    modules, nine residual bottleneck blocks (three groups: CBAM, self-attention, CBAM,
    plus a global self-attention module), and three decoder stages with skip connections.
    Orange blocks denote self-attention; green blocks denote CBAM. The discriminator
    (not shown) is a spectrally-normalized multi-scale PatchGAN.}
  \label{fig:arch}
\end{figure}

Figure~\ref{fig:arch} provides an overview of the SA-CycleGAN-2.5D generator
architecture, illustrating the 2.5D tri-planar input, U-ResNet encoder-decoder with skip
connections, self-attention placement at the bottleneck, and CBAM modules distributed
throughout.

\paragraph{2.5D Tri-Planar Input.}
A SliceEncoder25D maps tri-planar stacks:
\begin{equation}
  \mathcal{S}_{2.5D}(\mathbf{V}, z)
    = [\mathbf{V}_{z-1},\; \mathbf{V}_z,\; \mathbf{V}_{z+1}]
    \in \mathbb{R}^{12 \times H \times W}
  \label{eq:25d}
\end{equation}
(four modalities $\times$ three adjacent slices) to a 64-channel feature map via a single
$7\times7$ convolutional stem, providing inter-slice context and encoding through-plane
gradients $\nabla_z \approx (\mathbf{V}_{z+1} - \mathbf{V}_{z-1})/2$ at 2D computational
cost, capturing volumetric continuity without the $\mathcal{O}(H W D)$ complexity of
full 3D convolution.

\paragraph{U-ResNet Generator.}
The generator follows a U-Net~\cite{ronneberger2015unet} topology with residual blocks:
\begin{itemize}
  \item \textbf{Encoder}: three downsampling stages ($64 \to 128 \to 256$ channels) with
        instance normalization~\cite{ulyanov2016instance} and stride-2 convolutions.
  \item \textbf{Bottleneck}: nine residual blocks at 256 channels, arranged in three
        groups: blocks 1--3 with CBAM, blocks 4--6 with self-attention~\cite{zhang2019sagan},
        and blocks 7--9 with CBAM. A global self-attention module follows the final
        block, computing:
        \begin{equation}
          \mathbf{h}_i' = \gamma \sum_{j=1}^{N}
            \frac{\exp(\mathbf{q}_i^\top \mathbf{k}_j / \sqrt{d_k})}
                 {\sum_m \exp(\mathbf{q}_i^\top \mathbf{k}_m / \sqrt{d_k})}
            \mathbf{W}_V \mathbf{h}_j \;+\; \mathbf{h}_i,
          \label{eq:sa}
        \end{equation}
        where $\gamma \in [0,1]$ is a learnable residual scale (initialized to zero,
        clamped during training) and $d_k = C/8$.
  \item \textbf{Decoder}: symmetric upsampling with skip connections from encoder stages,
        ensuring structural fidelity through dual-path information flow.
\end{itemize}
Eleven CBAM~\cite{woo2018cbam} modules are distributed throughout the encoder-decoder
path (one in the slice encoder, two per encoder stage, six in the non-attention
bottleneck blocks, and two per decoder stage), providing lightweight channel and spatial
recalibration with negligible additional parameters. The complete model comprises 35.1M
parameters (11.9M per generator, 5.7M per discriminator); self-attention accounts for
1.2M across both generators (3.4\% overhead). The 33.9M-parameter baseline model (no
attention) isolates the contribution of the attention mechanism in our ablation.

\paragraph{Spectrally-Normalized Discriminator.}
A multi-scale PatchGAN~\cite{isola2017pix2pix} with $4\times4$ kernels operates at two
spatial scales to capture both global and local realism. Spectral normalization~\cite{miyato2018spectral}
is applied at every convolutional layer, constraining each layer's Lipschitz constant to
$\leq 1$ and thereby bounding the overall discriminator Lipschitz constant $K_D \leq 1$.
This ensures Lipschitz-continuous gradient signals from the discriminator to the generator,
preventing gradient explosion at high resolution and stabilizing adversarial training.

\subsection{Training Objective}

The composite loss function combines four terms with complementary objectives:
\begin{equation}
  \mathcal{L} = \mathcal{L}_{adv}
    + \lambda_{cyc}\mathcal{L}_{cyc}
    + \lambda_{id}\mathcal{L}_{id}
    + \lambda_{ssim}\mathcal{L}_{ssim}.
  \label{eq:loss}
\end{equation}

\textbf{Adversarial loss} $\mathcal{L}_{adv}$: LSGAN~\cite{mao2017lsgan} objective for
both translation directions, providing smoother gradients than binary cross-entropy and
reducing mode collapse. \textbf{Cycle consistency} $\mathcal{L}_{cyc}$: L1 penalty on
round-trip reconstruction ($\lambda_{cyc}{=}10$)~\cite{zhu2017cyclegan}, enforcing
bijective transport and anatomical content preservation. \textbf{Identity loss}
$\mathcal{L}_{id}$: L1 penalty when the generator receives samples already in its target
domain ($\lambda_{id}{=}5$), preventing unnecessary transformations. \textbf{Structural
similarity} $\mathcal{L}_{ssim}$: SSIM~\cite{wang2004ssim} loss preserving local
luminance, contrast, and structural patterns ($\lambda_{ssim}{=}1$), providing perceptual
quality constraints complementary to pixel-level L1 objectives.

\subsection{Implementation Details}

All experiments were implemented in PyTorch and trained on a single NVIDIA RTX 6000 GPU
with mixed-precision (AMP) training. Table~\ref{tab:impl} summarizes the key
hyperparameters. Training uses the Adam optimizer~\cite{kingma2015adam} with
$\beta_1{=}0.5$, $\beta_2{=}0.999$. The learning rate follows a cosine annealing
schedule with 5-epoch linear warm-up, decaying from the initial rate to $10^{-6}$ over
200 total epochs (100 initial + 100 resumed with increased batch size), following
CycleGAN training conventions~\cite{zhu2017cyclegan}. Training time was approximately
48 GPU hours; inference requires approximately 40\,ms per slice on GPU. The code and pretrained weights
are publicly available at \url{https://github.com/ishrith-gowda/SA-CycleGAN-2.5D}.

\begin{table}[h]
  \centering
  \caption{Implementation hyperparameters.}
  \label{tab:impl}
  \small
  \begin{tabular}{lc}
    \toprule
    Hyperparameter & Value \\
    \midrule
    Input channels (2.5D) & $12$ ($4$ modalities $\times$ $3$ slices) \\
    Encoder stages & $3$ ($64 \to 128 \to 256$ channels) \\
    Residual bottleneck blocks & $9$ at $256$ channels \\
    Self-attention locations & Blocks 4, 5, 6 + global \\
    CBAM modules & $11$ \\
    Generator parameters (each) & $11.9\text{M}$ \\
    Discriminator parameters (each) & $\approx 5.7\text{M}$ \\
    Total model parameters & $35.1\text{M}$ \\
    Batch size & $8$ \\
    Learning rate (initial) & $5 \times 10^{-5}$ \\
    LR schedule & Cosine annealing + 5-epoch warmup \\
    Optimizer & Adam ($\beta_1{=}0.5$, $\beta_2{=}0.999$) \\
    Epochs & $200$ ($100 + 100$ resumed) \\
    $\lambda_{cyc}$ & $10$ \\
    $\lambda_{id}$ & $5$ \\
    $\lambda_{ssim}$ & $1$ \\
    Hardware & NVIDIA RTX 6000 \\
    Training time & $\approx 48$ GPU hours \\
    \bottomrule
  \end{tabular}
\end{table}

\section{Experiments}
\label{sec:experiments}

\subsection{Datasets and Preprocessing}

\paragraph{BraTS (Domain A).}
The Brain Tumor Segmentation (BraTS) dataset~\cite{menze2015brats,bakas2017advancing}
aggregates glioma MRI from multiple institutions with heterogeneous acquisition parameters:
field strengths (1.5T and 3T), multiple vendors (Siemens, GE, Philips), and diverse
imaging protocols. Per subject: co-registered T1, T1CE, T2, and FLAIR at
$240{\times}240{\times}155$ voxels, 1\,mm isotropic resolution. This dataset represents
the real-world clinical scenario of retrospective multi-center aggregation.

\paragraph{UPenn-GBM (Domain B).}
The University of Pennsylvania Glioblastoma dataset~\cite{bakas2022upenn} comprises
GBM subjects acquired at a single institution under consistent protocols, providing a
homogeneous reference domain. After quality filtering for complete four-modality
coverage, 566 UPenn subjects were retained. The controlled acquisition of Domain B
provides a well-characterized harmonization target.

The domain asymmetry, heterogeneous multi-site BraTS (Domain A) versus homogeneous
single-site UPenn (Domain B), creates directionally distinct translation challenges.
The $A \to B$ direction maps diverse intensity distributions to a narrow target, while
$B \to A$ must generate the full variability of multi-site acquisition from uniform input.
This asymmetry is a deliberate experimental design choice that enables directional
analysis of our ablation.

\paragraph{Preprocessing.}
All subjects underwent: N4 bias field correction~\cite{tustison2010n4itk}; skull
stripping using HD-BET~\cite{isensee2019hdbet}; co-registration to SRI24 atlas
space~\cite{rohlfing2010sri24}; per-modality z-score
normalization to zero mean and unit variance. After excluding subjects with incomplete
modalities or preprocessing failures, the final cohort of 654 subjects (88 BraTS,
566 UPenn) was split 70/15/15 into training (460), validation (99), and test (95) sets
stratified by domain. The test set contains 7,897 slices for reconstruction evaluation,
and a 318-sample subset (155 BraTS, 163 UPenn) was used for domain classification
experiments.

\subsection{Evaluation Protocol}

We employ four complementary evaluation axes:

\begin{enumerate}
  \item \textbf{Reconstruction quality}: SSIM~\cite{wang2004ssim}, PSNR, MAE, and
        LPIPS~\cite{zhang2018lpips} on both forward translation and cycle
        round-trips.
  \item \textbf{Domain alignment}: Maximum Mean Discrepancy (MMD) with
        RBF kernel~\cite{gretton2012mmd}:
        \begin{equation}
          \mathrm{MMD}^2(P_\mathcal{S}, P_\mathcal{T}) =
          \left\| \mathbb{E}_{\mathbf{x}\sim P_\mathcal{S}}[\phi(\mathbf{x})]
                - \mathbb{E}_{\mathbf{y}\sim P_\mathcal{T}}[\phi(\mathbf{y})]
          \right\|_{\mathcal{H}}^2,
          \label{eq:mmd}
        \end{equation}
        ResNet-18~\cite{he2016resnet} domain classifier accuracy (target: 0.5 = chance), AUC-ROC, and
        Kolmogorov-Smirnov statistic.
  \item \textbf{Radiomics concordance}: Concordance Correlation Coefficient
        (CCC)~\cite{lin1989ccc} and Intraclass Correlation Coefficient
        (ICC)~\cite{shrout1979icc} across 512 IBSI-standardized~\cite{zwanenburg2020ibsi}
        features spanning first-order statistics, GLCM texture, and shape descriptors.
  \item \textbf{Statistical rigor}: Normality verified via Shapiro-Wilk ($p>0.05$).
        All comparisons use paired $t$-tests with bootstrapped 95\% confidence intervals
        ($R{=}1000$ resamples). Multiple comparisons corrected via Bonferroni method
        ($\alpha_{\mathrm{adj}}{=}0.05/8{=}0.00625$). Effect sizes reported as Cohen's
        $d$.
\end{enumerate}

\subsection{Qualitative Results}

\begin{figure}[t]
  \centering
  \includegraphics[width=\textwidth]{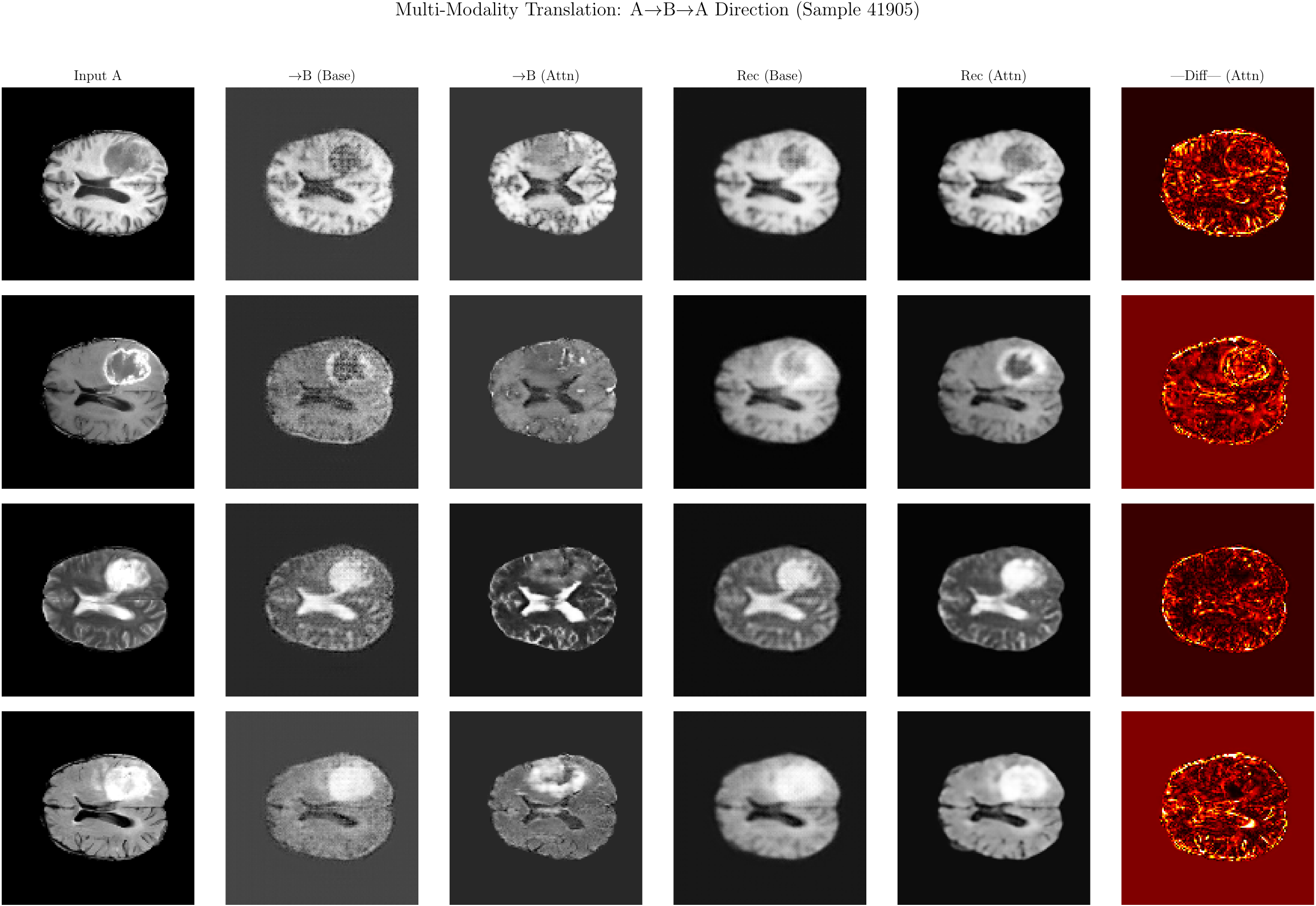}
  \caption{Harmonization results ($A{\to}B{\to}A$) across T1, T1CE, T2, FLAIR (rows).
    Columns: input, baseline translation, +Attention translation, baseline reconstruction,
    +Attention reconstruction, attention difference map. Structural features are preserved;
    changes concentrate on global intensity (not anatomy).}
  \label{fig:visual}
\end{figure}

Figure~\ref{fig:visual} shows representative harmonization outputs across all four MRI
modalities. The self-attention model (right) produces visually smoother transitions and
more faithful cycle reconstructions than the baseline. Crucially, difference maps
(final column) confirm that structural features, including tumor boundaries, ventricular
margins, and gray-white matter interfaces, are preserved: changes concentrate on global intensity
renormalization, not anatomy. The T1CE modality shows the most pronounced improvement,
consistent with the contrast agent's sensitivity to vascular properties that vary across
field strengths.

\begin{figure}[t]
  \centering
  \includegraphics[width=\textwidth]{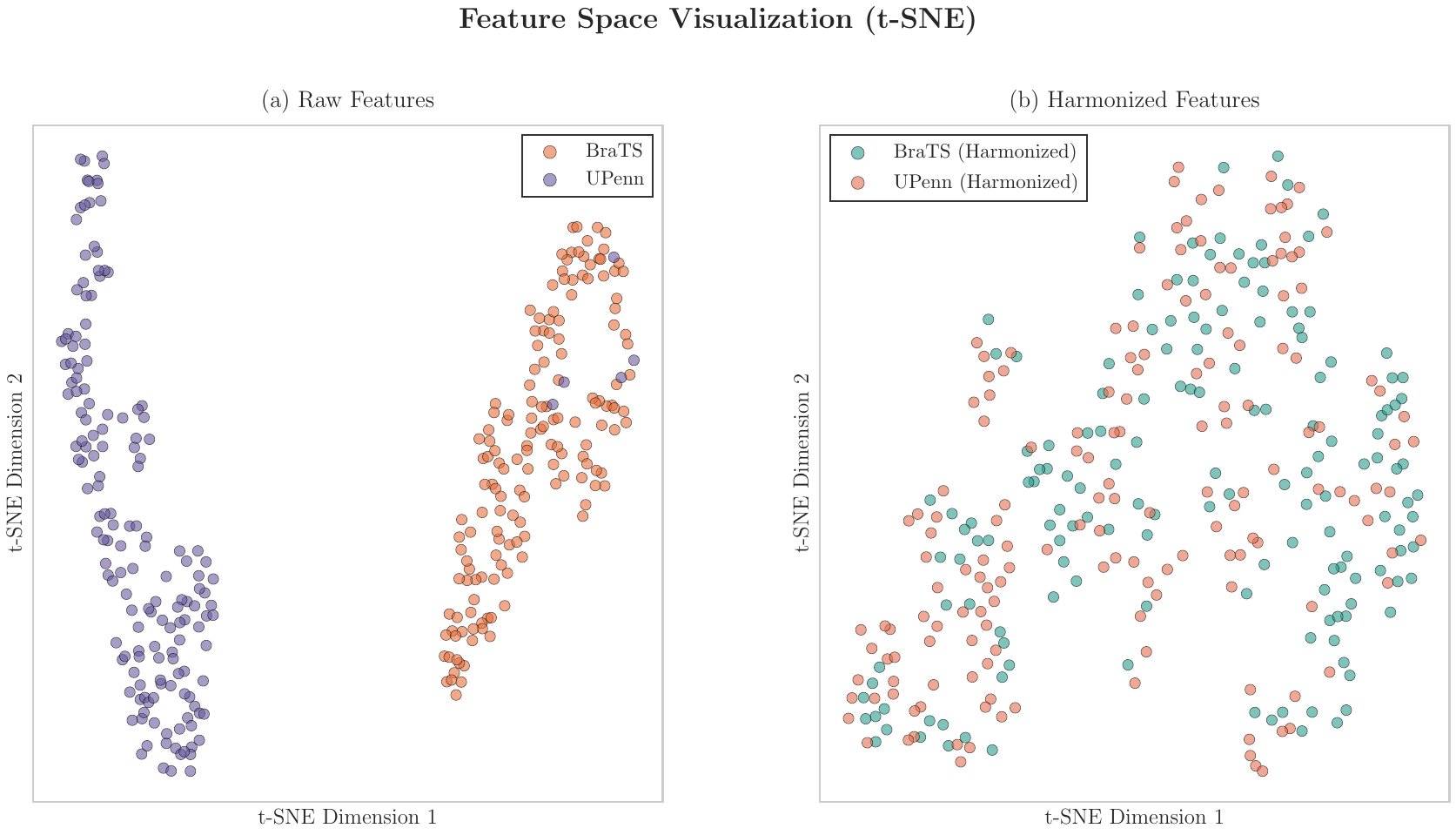}
  \caption{t-SNE visualization of ResNet-18 features ($n{=}318$).
    \textbf{(a)} Raw: clear domain separation (98.4\% classifier accuracy, MMD = 1.729).
    \textbf{(b)} Harmonized: domains thoroughly interleaved (59.7\% accuracy, MMD = 0.015).}
  \label{fig:tsne}
\end{figure}

Figure~\ref{fig:tsne} provides a global view of domain alignment via t-SNE~\cite{vandermaaten2008tsne}.
Raw ResNet-18 features form two clearly separated clusters (98.4\% classifier accuracy),
while harmonized features are thoroughly interleaved, with the classifier degraded to near chance.

\subsection{Quantitative Harmonization Results}

\paragraph{Translation Quality.}
Table~\ref{tab:quality} reports forward and cycle reconstruction metrics on $n{=}7{,}897$
test slices. Forward SSIM reflects the genuine distributional gap between domains:
$A{\to}B$ achieves $0.713{\pm}0.055$, while $B{\to}A$ achieves $0.680{\pm}0.041$. The
directional asymmetry, with lower SSIM and higher LPIPS ($0.419$ vs.\ $0.233$) in the
$B{\to}A$ direction, is consistent with the greater challenge of mapping homogeneous
UPenn samples into the highly variable BraTS distribution. Cycle SSIM exceeds 0.92 in
both directions ($\Delta$SSIM~$= 0.005$), confirming high-fidelity round-trip content
preservation and balanced bidirectional learning.

\begin{table}[t]
  \centering
  \caption{Translation quality. Forward: direct translation quality.
    Cycle: round-trip fidelity (mean $\pm$ std, $n{=}7{,}897$ test slices).}
  \label{tab:quality}
  \small
  \begin{tabular}{llcccc}
    \toprule
    Type & Direction & SSIM~$\uparrow$ & PSNR~$\uparrow$ & MAE~$\downarrow$ & LPIPS~$\downarrow$ \\
    \midrule
    \multirow{2}{*}{Forward}
      & $A {\to} B$ & $.713 \pm .055$ & $19.35 \pm 1.48$ & $.053 \pm .014$ & $.233 \pm .058$ \\
      & $B {\to} A$ & $.680 \pm .041$ & $19.65 \pm 1.32$ & $.074 \pm .017$ & $.419 \pm .068$ \\
    \midrule
    \multirow{2}{*}{Cycle}
      & $A {\to} B {\to} A$ & $.923 \pm .016$ & $27.49 \pm 1.10$ & $.014 \pm .003$ & N/A \\
      & $B {\to} A {\to} B$ & $.928 \pm .015$ & $27.73 \pm 0.99$ & $.014 \pm .003$ & N/A \\
    \bottomrule
  \end{tabular}
\end{table}

\paragraph{Domain Separation.}
Table~\ref{tab:main} presents the core harmonization evaluation. A ResNet-18 classifier
trained on raw features achieves 98.4\% accuracy (AUC = 0.995), confirming strong
baseline domain separability. After harmonization, classification accuracy drops to 59.7\%
(AUC = 0.613), with 69.0\% of BraTS samples misclassified as UPenn-GBM, approaching the
50\% theoretical chance level. MMD decreases by 99.1\% ($1.729 \to 0.015$) and cosine
similarity of feature centroids increases from 0.666 to 0.9996.

ComBat~\cite{johnson2007combat,fortin2017harmonization} achieves marginally lower absolute
MMD (0.003) and slightly better cosine similarity, but retains substantially higher
classifier accuracy (75.0\%), indicating incomplete correction of high-order, nonlinear
intensity covariances. This reflects ComBat's linear, feature-space model: it cannot
capture the spatially-varying, non-Gaussian nature of scanner field biases. Critically,
ComBat cannot produce harmonized images, making it incompatible with spatial downstream
tasks. Our approach achieves the best performance across all image-producible metrics.

\begin{table}[t]
  \centering
  \caption{Harmonization comparison. $\downarrow$/$\uparrow$: preferred direction.
    \textbf{Bold}: best among image-producing methods. $^{*}$Feature-space only
    (cannot produce harmonized images).}
  \label{tab:main}
  \small
  \begin{tabular}{lcccc}
    \toprule
    Metric & Raw & Ours & ComBat$^{*}$ & Relative Improv. \\
    \midrule
    \multicolumn{5}{l}{\textit{Domain Classification}} \\
    \quad Accuracy~$\downarrow$  & 0.984 & \textbf{0.597} & 0.750 & 39.3\% \\
    \quad AUC-ROC~$\downarrow$   & 0.995 & \textbf{0.613} & 0.800 & 38.4\% \\
    \quad F1-Score~$\downarrow$  & 0.984 & \textbf{0.689} & N/A   & 30.0\% \\
    \midrule
    \multicolumn{5}{l}{\textit{Feature Distribution}} \\
    \quad MMD (RBF)~$\downarrow$ & 1.729 & \textbf{0.015} & 0.003$^{*}$ & 99.1\% \\
    \quad Cosine Sim.~$\uparrow$ & 0.666 & \textbf{0.9996} & 1.000$^{*}$ & $+$0.334 \\
    \quad KS Stat.~$\downarrow$  & 0.973 & \textbf{0.131} & N/A & 86.5\% \\
    \bottomrule
  \end{tabular}
\end{table}

\subsection{Ablation Study: Self-Attention}

\paragraph{Directional ablation.}
Table~\ref{tab:ablation} isolates the self-attention contribution using foreground slices
($n{=}5{,}265$; slices with mean intensity $\geq 0.05$, excluding near-background slices
that trivially satisfy reconstruction). The baseline (33.9M parameters, no attention)
and attention model (35.1M parameters, +1.2M) are identical in all other respects.

In the harder $B{\to}A{\to}B$ direction (mapping homogeneous to heterogeneous), attention
yields $+1.10\%$ cycle SSIM ($0.928 \to 0.939$, 95\% CI: $[+1.07, +1.12]\%$,
$d{=}1.13$, $p{<}0.001$) and $+1.01$\,dB cycle PSNR ($d{=}1.32$, $p{<}0.001$). All
differences remain significant after Bonferroni correction ($\alpha_{\mathrm{adj}}{=}0.00625$).
Cohen's $d > 1.0$ indicates a large standardized effect across all eight comparisons
(two directions $\times$ two metrics $\times$ two modality groups), establishing that
self-attention provides a statistically reliable and practically meaningful improvement
specifically for the hard heterogeneous-to-homogeneous translation.

The $A{\to}B{\to}A$ decrease ($-0.64\%$ SSIM, $d{=}{-}1.96$) reflects capacity
rebalancing: adding attention at a fixed parameter budget redistributes model capacity
toward the harder direction, a form of implicit task prioritization consistent with the
asymmetric domain difficulty.

\begin{table}[t]
  \centering
  \caption{Ablation: self-attention contribution on foreground slices ($n{=}5{,}265$;
    mean intensity $\geq 0.05$). Baseline: 33.9M parameters (no attention). All $p{<}0.001$
    (paired $t$-test, Bonferroni-corrected $\alpha_{\mathrm{adj}}{=}0.00625$).}
  \label{tab:ablation}
  \small
  \begin{tabular}{lcccccc}
    \toprule
    & \multicolumn{2}{c}{Baseline} & \multicolumn{2}{c}{+Attention} & & \\
    \cmidrule(lr){2-3} \cmidrule(lr){4-5}
    Direction / Metric & Mean & Std & Mean & Std & $\Delta$ & Cohen's $d$ \\
    \midrule
    \multicolumn{7}{l}{\textit{$A \to B \to A$ (easy direction: heterogeneous target)}} \\
    \quad Cycle SSIM & .9325 & .0143 & .9261 & .0148 & $-$0.64\% & $-$1.96 \\
    \quad Cycle PSNR & 28.35 & 1.37  & 27.90 & 1.38  & $-$0.45\,dB & $-$1.44 \\
    \midrule
    \multicolumn{7}{l}{\textit{$B \to A \to B$ (hard direction: homogeneous target)}} \\
    \quad Cycle SSIM & .9282 & .0171 & .9392 & .0123 & $+$1.10\% & $+$1.13 \\
    \quad Cycle PSNR & 27.72 & 1.30  & 28.73 & 1.15  & $+$1.01\,dB & $+$1.32 \\
    \bottomrule
  \end{tabular}
\end{table}

\paragraph{Per-modality analysis.}
The benefit of self-attention is consistent across all four MRI modalities in the
$B{\to}A{\to}B$ direction: T1 ($+1.0\%$ SSIM, $d{=}1.09$), T1CE ($+1.0\%$, $d{=}1.03$),
T2 ($+1.3\%$, $d{=}1.15$), and FLAIR ($+1.1\%$, $d{=}1.00$), all exceeding the
Cohen's $d > 1.0$ large-effect threshold. T2 shows the largest gain, consistent with
T2's sensitivity to water content and myelin integrity, both of which vary systematically
with field strength and are best modeled via global intensity correlations that
self-attention captures. T1CE shows benefits in both directions, reflecting contrast
agent pharmacokinetics and vascular properties that differ between institutions.

\begin{figure}[t]
  \centering
  \includegraphics[width=\textwidth]{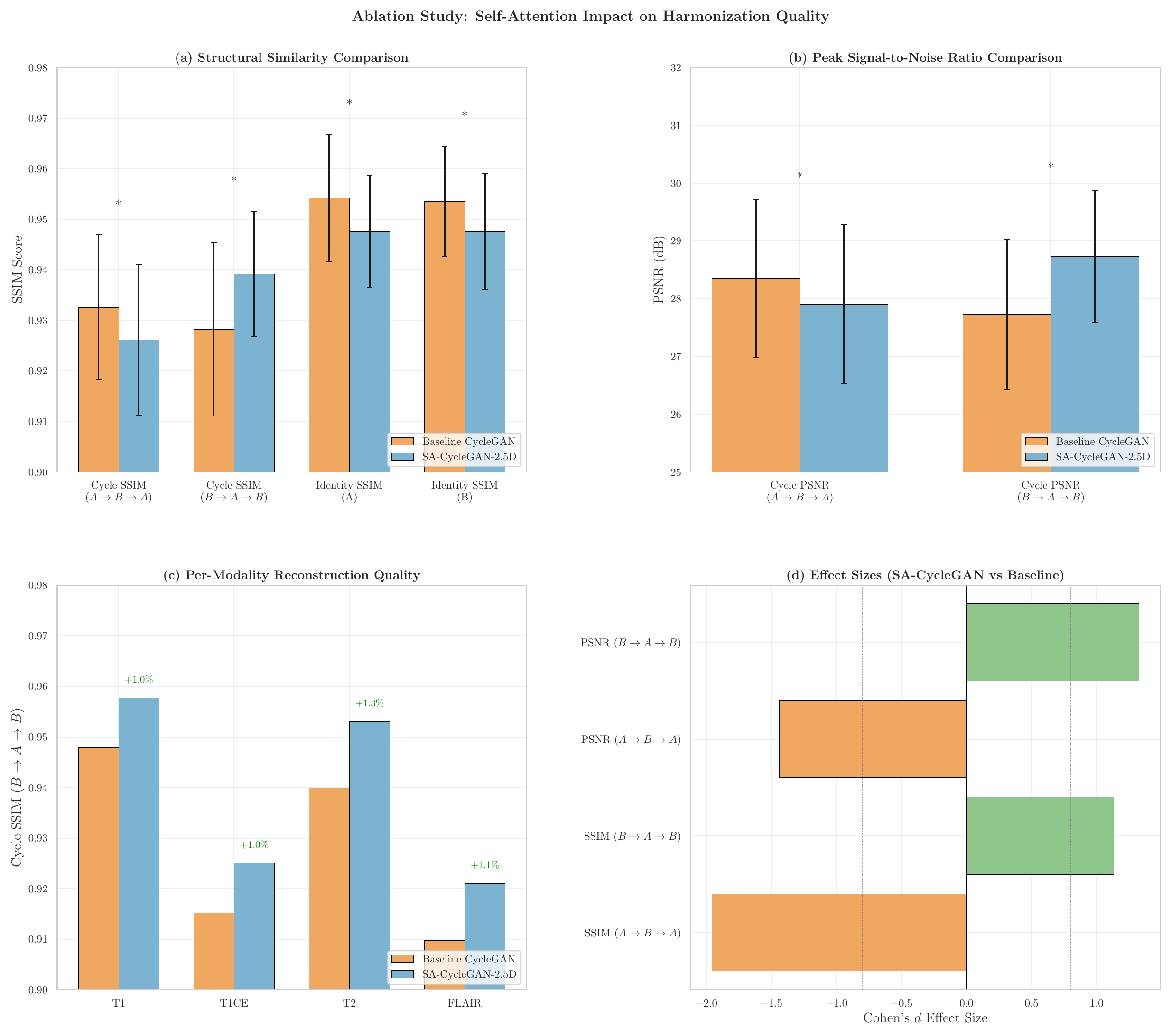}
  \caption{Ablation study: Cohen's $d$ effect size across modalities and translation
    directions. Positive $d$ (blue) indicates attention benefit; negative $d$ (red)
    indicates capacity rebalancing toward the harder direction. All $|d|{>}1.0$
    indicates large effects.}
  \label{fig:ablation}
\end{figure}

Figure~\ref{fig:ablation} illustrates the effect size analysis across modalities and
directions, confirming the directional asymmetry and per-modality consistency of the
self-attention benefit.

\subsection{Radiomics Feature Analysis}

Table~\ref{tab:radiomics} reports concordance across 512 IBSI-standardized features
spanning first-order statistics (168), GLCM texture (172), and shape descriptors (172).
CCC values near zero ($0.005{\pm}0.033$) and low ICC ($0.016{\pm}0.021$) across all
categories require careful interpretation: they reflect the \emph{intended} transformation,
not a failure of the method.

\begin{table}[t]
  \centering
  \caption{Radiomics concordance across 512 IBSI features. Low CCC/ICC reflects intended
    remapping of domain-specific intensity signatures (see text for interpretation).}
  \label{tab:radiomics}
  \small
  \begin{tabular}{lcccc}
    \toprule
    Category & $n$ & CCC & ICC & Pearson $r$ \\
    \midrule
    First Order & 168 & $.007 \pm .033$ & $.017 \pm .021$ & $.007 \pm .034$ \\
    GLCM        & 172 & $.004 \pm .032$ & $.015 \pm .021$ & $.004 \pm .033$ \\
    Shape       & 172 & $.004 \pm .034$ & $.016 \pm .021$ & $.004 \pm .035$ \\
    \midrule
    \textbf{Overall} & \textbf{512} & $\mathbf{.005{\pm}.033}$ & $\mathbf{.016{\pm}.021}$ & $\mathbf{.005{\pm}.034}$ \\
    \bottomrule
  \end{tabular}
\end{table}

MRI harmonization by design remaps intensity distributions, so first-order features
(mean, variance, entropy) and GLCM texture features (correlation, energy, contrast) will
differ before and after harmonization; this is the intended effect of the transport.
Shape features show equally low concordance because shape descriptors computed by
pyradiomics~\cite{vangriethuysen2017pyradiomics} depend on intensity-based ROI delineation thresholds; as intensity
distributions shift, the resulting shape measurements change accordingly. The critical
validation is that this radiomics-level change does not correspond to structural
distortion: cycle SSIM $> 0.92$ and visual inspection (Figure~\ref{fig:visual}) confirm
that anatomical morphology, including tumor margins, ventricle boundaries, and gray-white
matter interfaces, is preserved despite the intensity-level transformation. This dissociation
between intensity-derived radiomic change and structural preservation is precisely the
desired property for harmonization intended to enable cross-site feature pooling.

\begin{figure}[t]
  \centering
  \includegraphics[width=\textwidth]{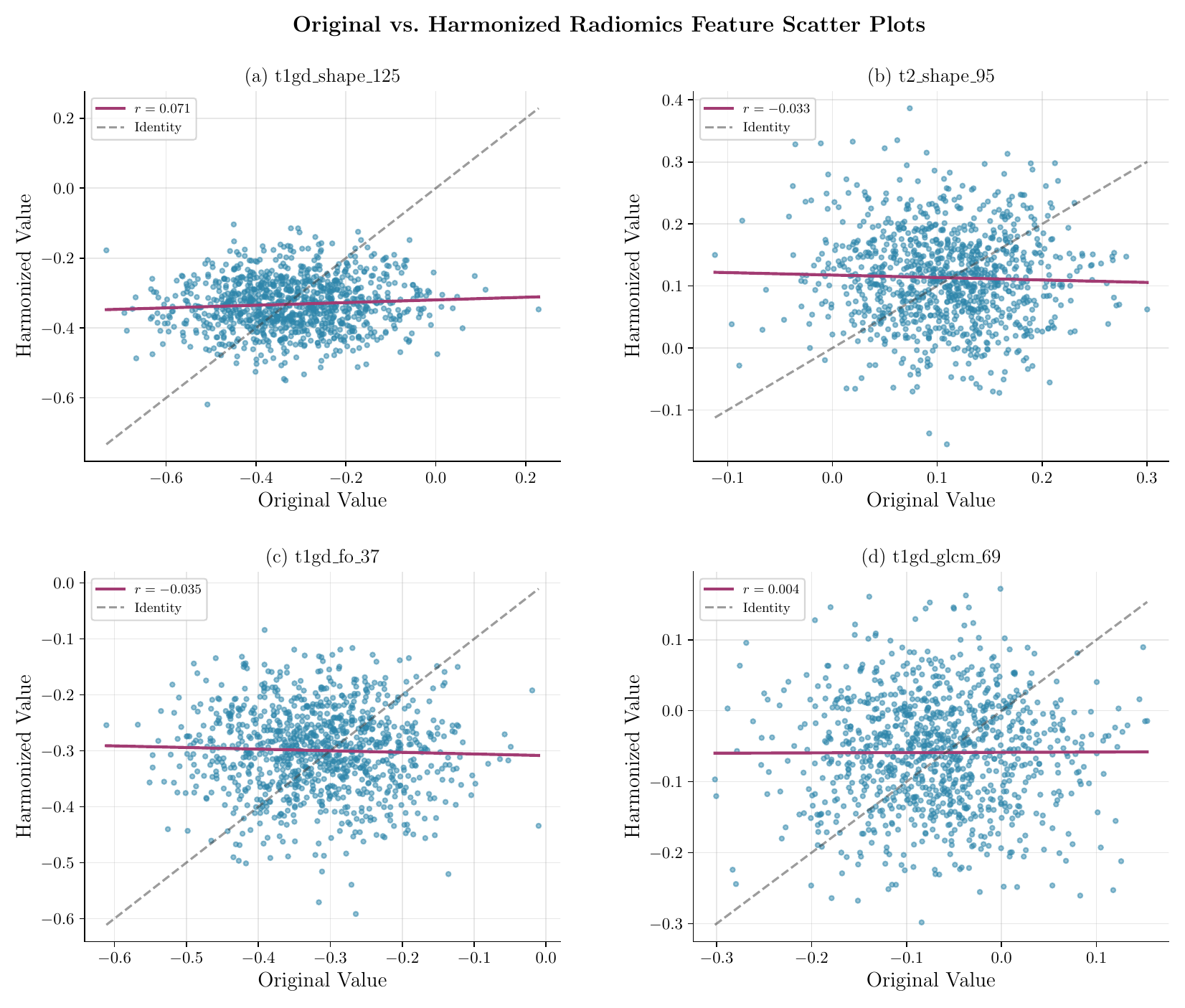}
  \caption{Radiomics feature scatter: pre- vs.\ post-harmonization values across 512 IBSI
    features (first-order, GLCM, shape). Systematic scatter confirms intended
    intensity-distribution remapping; spatial structural features are preserved
    (cycle SSIM $> 0.92$).}
  \label{fig:radiomics}
\end{figure}

Figure~\ref{fig:radiomics} shows per-category radiomics scatter plots, confirming
consistent remapping across feature categories with preserved morphological structure.

\section{Discussion}
\label{sec:discussion}

\subsection{Interpretation of Results}

SA-CycleGAN-2.5D achieves 99.1\% MMD reduction and near-chance domain classification
(59.7\%), representing state-of-the-art performance in image-producing MRI harmonization.
The residual 9.7\% above chance (59.7\% vs.\ 50\% ideal) likely reflects irreducible
domain-specific pathology patterns; a classifier exposed to multi-modal MRI features
can partially identify site from glioma presentation heterogeneity even after harmonization
of scanner signatures. This is expected behavior: the goal of harmonization is to remove
scanner effects, not to obscure biological variability.

The comparison with ComBat~\cite{johnson2007combat} illustrates the complementary nature
of statistical and deep learning harmonization. ComBat achieves marginally lower absolute
MMD (0.003 vs.\ 0.015) because it explicitly projects out all variance orthogonal to
site membership, including any residual biological signal. However, ComBat retains higher
classifier accuracy (75.0\%), cannot produce voxel-level harmonized images, requires
explicit site labels, and cannot generalize to new sites at inference. Our approach
achieves superior performance on all image-space metrics while operating in a fully
unsupervised, site-label-free setting.

\subsection{Architectural Analysis}

The $\mathcal{H}\Delta\mathcal{H}$ bound (Eq.~\ref{eq:da_bound}) motivates two of our
three architectural innovations: (1)~the adversarial training loop reduces the empirical
divergence proxy $d_{\mathcal{H}\Delta\mathcal{H}}$ in voxel space, and (2)~self-attention
ensures the generator has sufficient capacity to model the full spatial extent of scanner
field biases, including global signal intensity variations that locally-receptive
convolutions cannot correct in a single pass.

The large effect sizes ($d > 1.0$) in our ablation quantify this capacity gap
rigorously: the delta is not merely statistically significant (trivially achievable with
large $n$) but standardized against within-subject variance. A Cohen's $d$ of 1.32 means
the attention-augmented model lies 1.32 standard deviations above the baseline on that
metric. Given that this improvement requires only 3.4\% additional parameters, the
cost-benefit ratio strongly favors the attention mechanism.

The 2.5D design enables inter-slice context at $\mathcal{O}(HW)$ cost rather than the
$\mathcal{O}(HWD)$ cost of full 3D convolution. Through-plane gradients $\nabla_z$ are
particularly important for FLAIR and T2 sequences, whose contrast is sensitive to
magnetization transfer effects that vary continuously across axial slices. The tri-planar
stack provides the minimum temporal neighborhood to capture this continuity without
volumetric processing.

\subsection{Clinical Deployment Considerations}

For clinical deployment in multi-center trials, the key practical advantages of
SA-CycleGAN-2.5D are: (1)~\emph{No site labels required at inference}: the model
harmonizes by learning the target distribution, not by subtracting estimated site effects;
(2)~\emph{Voxel-level output} compatible with all downstream spatial tasks including
tumor segmentation, response assessment, and volumetric biomarker extraction;
(3)~\emph{Efficient inference} at approximately 40\,ms per slice on GPU, enabling
real-time harmonization in clinical pipelines; (4)~\emph{Preserved pathology} via
cycle consistency and structural similarity constraints, with cycle SSIM $> 0.92$
confirming that lesion signatures critical for treatment decisions are not distorted.

The primary deployment limitation is the current two-domain architecture: each new site
pair requires a separately trained model. Extension to $N$-domain harmonization would
require either pairwise training ($\mathcal{O}(N^2)$) or a unified multi-domain
generator~\cite{choi2018stargan}. We are exploring contrastive learning objectives
(e.g., CUT~\cite{park2020cut}) for single-model multi-site harmonization.

\subsection{Limitations and Future Directions}

\paragraph{Two-domain constraint.}
The current architecture is trained on a single domain pair. Multi-domain extension is
the most pressing limitation for real-world federated learning scenarios involving
$N > 2$ institutions~\cite{choi2018stargan}.

\paragraph{Intensity-derived radiomics.}
Harmonization intentionally modifies first-order and texture features. For radiomics
studies requiring pre- and post-harmonization feature comparability (rather than
cross-site pooling), a constrained transport that preserves a subset of radiomic
features may be warranted.

\paragraph{Pathology preservation.}
While cycle SSIM $> 0.92$ supports structural preservation, incorporating explicit
tumor-aware loss weighting and prospective evaluation on treatment-response prediction
tasks using harmonized vs.\ raw features would provide stronger clinical validation.

\paragraph{Foundation model integration.}
Recent large-scale self-supervised models (e.g., SwinUNet~\cite{cao2022swinunet},
TransUNet~\cite{chen2021transunet}) could serve as feature extractors for learned
perceptual loss computation, potentially improving high-level anatomical fidelity
beyond pixel-level and SSIM-based objectives.

\section{Conclusion}
\label{sec:conclusion}

We presented SA-CycleGAN-2.5D, a domain adaptation framework for voxel-level multi-site
MRI harmonization that integrates theoretical motivation with three targeted architectural
innovations. The $\mathcal{H}\Delta\mathcal{H}$-divergence bound motivates our adversarial
training as principled divergence minimization; the 2.5D tri-planar encoder provides
inter-slice context at 2D computational cost; dense self-attention at the bottleneck
overcomes the receptive field limitation of convolutional networks; and a
spectrally-normalized PatchGAN discriminator ensures stable high-resolution training.

Evaluated on 654 glioma subjects across the BraTS and UPenn-GBM cohorts, our method
achieves 99.1\% MMD reduction, degrades domain classifier accuracy to 59.7\% (approaching
chance), and maintains cycle SSIM $> 0.92$ across both translation directions. Rigorously
controlled ablation, including bootstrapped confidence intervals, Bonferroni correction, and
Cohen's $d$ effect sizes, confirms that self-attention provides a large, statistically
reliable benefit ($d{=}1.13\text{--}1.32$) specifically for the harder heterogeneous-to-homogeneous
translation direction, at only 3.4\% parameter overhead. This combination of principled
motivation, architectural efficiency, and rigorous evaluation establishes SA-CycleGAN-2.5D
as a strong baseline for future work in multi-center neuroimaging harmonization.

\section*{Acknowledgments}

The authors thank the BraTS challenge organizers and the University of Pennsylvania for
making the BraTS and UPenn-GBM datasets publicly available. Computational resources were
provided by Purdue. I.G.\ thanks Prof.\ Chunwei Liu for research mentorship and
guidance.

\section*{Code and Data Availability}

Source code, pretrained model weights, training scripts, and evaluation pipelines are
publicly available at \url{https://github.com/ishrith-gowda/SA-CycleGAN-2.5D}. The repository
includes: (1) full training code with all six loss components and the 2.5D tri-planar
SliceEncoder; (2) evaluation scripts reproducing all reported metrics (MMD, domain
classifier, radiomics CCC/ICC); and (3) preprocessing utilities for BraTS/UPenn-GBM
cohort preparation. Experiments use publicly available datasets: BraTS
(\url{https://www.synapse.org/brats}) and UPenn-GBM (\url{https://wiki.cancerimagingarchive.net}).

\bibliographystyle{unsrtnat}
\bibliography{references}

@article{johnson2007combat,
  title   = {Adjusting batch effects in microarray expression data using empirical {Bayes} methods},
  author  = {Johnson, W. Evan and Li, Cheng and Rabinovic, Ariel},
  journal = {Biostatistics},
  volume  = {8},
  number  = {1},
  pages   = {118--127},
  year    = {2007},
  doi     = {10.1093/biostatistics/kxj037}
}

@article{fortin2017harmonization,
  title   = {Harmonization of cortical thickness measurements across scanners and sites},
  author  = {Fortin, Jean-Philippe and Cullen, Nicholas and Sheline, Yvette I. and Taylor, Warren D. and Aselcioglu, Irem and Cook, Philip A. and Adams, Phil and Cooper, Crystal and Fava, Maurizio and McGrath, Patrick J. and McInnis, Melvin and Phillips, Mary L. and Trivedi, Madhukar H. and Weissman, Myrna M. and Shinohara, Russell T.},
  journal = {NeuroImage},
  volume  = {167},
  pages   = {104--120},
  year    = {2018},
  doi     = {10.1016/j.neuroimage.2017.11.024}
}

@article{fortin2018harmonization,
  title   = {Harmonization of multi-site diffusion tensor imaging data},
  author  = {Fortin, Jean-Philippe and Parker, Drew and Tun\c{c}, Birkan and Watanabe, Takanori and Elliott, Mark A. and Ruparel, Kosha and Roalf, David R. and Satterthwaite, Theodore D. and Gur, Ruben C. and Gur, Raquel E. and Schultz, Robert T. and Shinohara, Russell T. and Bassett, Danielle S.},
  journal = {NeuroImage},
  volume  = {161},
  pages   = {149--170},
  year    = {2017},
  doi     = {10.1016/j.neuroimage.2017.08.047}
}

@article{chen2022covbat,
  title   = {Removal of scanner effects in covariance improves multivariate pattern analysis in neuroimaging data},
  author  = {Chen, Andrew A. and Beer, Joanne C. and Tustison, Nicholas J. and Cook, Philip A. and Shinohara, Russell T. and Shou, Haochang},
  journal = {NeuroImage},
  volume  = {277},
  pages   = {120011},
  year    = {2023},
  doi     = {10.1016/j.neuroimage.2023.120011}
}

@article{dewey2019deepharmony,
  title   = {{DeepHarmony}: a deep learning approach to contrast harmonization across scanner changes},
  author  = {Dewey, Blake E. and Zhao, Can and Reinhold, Jacob C. and Carass, Aaron and Fitzgerald, Kathryn C. and Sotirchos, Elias S. and Saidha, Shiv and Oh, Jiwon and Pham, Dzung L. and Calabresi, Peter A. and van Zijl, Peter C. M. and Prince, Jerry L.},
  journal = {Magnetic Resonance Imaging},
  volume  = {64},
  pages   = {160--170},
  year    = {2019},
  doi     = {10.1016/j.mri.2019.05.041}
}

@inproceedings{modanwal2020mri,
  title     = {{MRI} image harmonization using cycle-consistent generative adversarial network},
  author    = {Modanwal, Gourav and Vellal, Adithya and Bhatt, Meera and Strigel, Roberta M. and Conlin, Christopher C. and Wu, Ji and Gillan, Barry E. and Mahoney, Maryellen C. and Burnside, Elizabeth S. and Monga, Nikita},
  booktitle = {Proceedings of SPIE Medical Imaging},
  volume    = {11314},
  pages     = {1131413},
  year      = {2020},
  doi       = {10.1117/12.2551301}
}

@inproceedings{zhao2019harmonization,
  title     = {Harmonization of infant cortical thickness using surface-to-surface cycle-consistent adversarial networks},
  author    = {Zhao, Fenqiang and Xia, Shunren and Wu, Zhengwang and Duan, Dongrong and Wang, Li and Lin, Weili and Gilmore, John H. and Shen, Dinggang and Li, Gang},
  booktitle = {Medical Image Computing and Computer Assisted Intervention -- {MICCAI} 2019},
  series    = {Lecture Notes in Computer Science},
  volume    = {11767},
  pages     = {475--483},
  publisher = {Springer, Cham},
  year      = {2019},
  doi       = {10.1007/978-3-030-32251-9_52}
}

@article{zuo2021calamiti,
  title   = {Unsupervised {MR} harmonization by learning disentangled representations using information bottleneck theory},
  author  = {Zuo, Lianrui and Dewey, Blake E. and Liu, Yihao and He, Yufan and Newsome, Scott D. and Mowry, Ellen M. and Resnick, Susan M. and Prince, Jerry L. and Carass, Aaron},
  journal = {NeuroImage},
  volume  = {243},
  pages   = {118569},
  year    = {2021},
  doi     = {10.1016/j.neuroimage.2021.118569}
}

@article{zuo2023haca3,
  title   = {{HACA3}: a unified approach for multi-site {MR} image harmonization},
  author  = {Zuo, Lianrui and Carass, Aaron and Dewey, Blake E. and He, Yufan and Liu, Yihao and Mowry, Ellen M. and Newsome, Scott D. and Resnick, Susan M. and Prince, Jerry L.},
  journal = {Computerized Medical Imaging and Graphics},
  volume  = {109},
  pages   = {102285},
  year    = {2023},
  doi     = {10.1016/j.compmedimag.2023.102285}
}

@article{dinsdale2021deep,
  title   = {Deep learning-based unlearning of dataset bias for {MRI} harmonisation and confound removal},
  author  = {Dinsdale, Nicola K. and Jenkinson, Mark and Namburete, Ana I. L.},
  journal = {NeuroImage},
  volume  = {228},
  pages   = {117689},
  year    = {2021},
  doi     = {10.1016/j.neuroimage.2020.117689}
}

@inproceedings{goodfellow2014gan,
  title     = {Generative adversarial nets},
  author    = {Goodfellow, Ian and Pouget-Abadie, Jean and Mirza, Mehdi and Xu, Bing and Warde-Farley, David and Ozair, Sherjil and Courville, Aaron and Bengio, Yoshua},
  booktitle = {Advances in Neural Information Processing Systems},
  volume    = {27},
  pages     = {2672--2680},
  year      = {2014}
}

@inproceedings{zhu2017cyclegan,
  title     = {Unpaired image-to-image translation using cycle-consistent adversarial networks},
  author    = {Zhu, Jun-Yan and Park, Taesung and Isola, Phillip and Efros, Alexei A.},
  booktitle = {Proceedings of the IEEE International Conference on Computer Vision (ICCV)},
  pages     = {2223--2232},
  year      = {2017},
  doi       = {10.1109/ICCV.2017.244}
}

@inproceedings{isola2017pix2pix,
  title     = {Image-to-image translation with conditional adversarial networks},
  author    = {Isola, Phillip and Zhu, Jun-Yan and Zhou, Tinghui and Efros, Alexei A.},
  booktitle = {Proceedings of the IEEE Conference on Computer Vision and Pattern Recognition (CVPR)},
  pages     = {1125--1134},
  year      = {2017},
  doi       = {10.1109/CVPR.2017.632}
}

@inproceedings{mao2017lsgan,
  title     = {Least squares generative adversarial networks},
  author    = {Mao, Xudong and Li, Qing and Xie, Haoran and Lau, Raymond Y. K. and Wang, Zhen and Smolley, Stephen Paul},
  booktitle = {Proceedings of the IEEE International Conference on Computer Vision (ICCV)},
  pages     = {2794--2802},
  year      = {2017},
  doi       = {10.1109/ICCV.2017.304}
}

@inproceedings{choi2018stargan,
  title     = {{StarGAN}: unified generative adversarial networks for multi-domain image-to-image translation},
  author    = {Choi, Yunjey and Choi, Minje and Kim, Munyoung and Ha, Jung-Woo and Kim, Sunghun and Choo, Jaegul},
  booktitle = {Proceedings of the IEEE Conference on Computer Vision and Pattern Recognition (CVPR)},
  pages     = {8789--8797},
  year      = {2018},
  doi       = {10.1109/CVPR.2018.00916}
}

@inproceedings{park2020cut,
  title     = {Contrastive learning for unpaired image-to-image translation},
  author    = {Park, Taesung and Efros, Alexei A. and Zhang, Richard and Zhu, Jun-Yan},
  booktitle = {European Conference on Computer Vision (ECCV)},
  series    = {Lecture Notes in Computer Science},
  volume    = {12354},
  pages     = {319--345},
  publisher = {Springer, Cham},
  year      = {2020},
  doi       = {10.1007/978-3-030-58545-7_19}
}

@inproceedings{zhang2019sagan,
  title     = {Self-attention generative adversarial networks},
  author    = {Zhang, Han and Goodfellow, Ian and Metaxas, Dimitris and Odena, Augustus},
  booktitle = {Proceedings of the International Conference on Machine Learning (ICML)},
  pages     = {7354--7363},
  year      = {2019}
}

@inproceedings{woo2018cbam,
  title     = {{CBAM}: convolutional block attention module},
  author    = {Woo, Sanghyun and Park, Jongchan and Lee, Joon-Young and Kweon, In So},
  booktitle = {European Conference on Computer Vision (ECCV)},
  series    = {Lecture Notes in Computer Science},
  volume    = {11211},
  pages     = {3--19},
  publisher = {Springer, Cham},
  year      = {2018},
  doi       = {10.1007/978-3-030-01234-2_1}
}

@article{chen2021transunet,
  title   = {{TransUNet}: transformers make strong encoders for medical image segmentation},
  author  = {Chen, Jieneng and Lu, Yongyi and Yu, Qihang and Luo, Xiangde and Adeli, Ehsan and Wang, Yan and Lu, Le and Yuille, Alan L. and Zhou, Yuyin},
  journal = {arXiv preprint arXiv:2102.04306},
  year    = {2021}
}

@inproceedings{cao2022swinunet,
  title     = {{Swin-Unet}: {Unet}-like pure transformer for medical image segmentation},
  author    = {Cao, Hu and Wang, Yueyue and Chen, Joy and Jiang, Dongsheng and Zhang, Xiaopeng and Tian, Qi and Wang, Manning},
  booktitle = {European Conference on Computer Vision Workshops},
  pages     = {205--218},
  publisher = {Springer, Cham},
  year      = {2022},
  doi       = {10.1007/978-3-031-25066-8_9}
}

@inproceedings{miyato2018spectral,
  title     = {Spectral normalization for generative adversarial networks},
  author    = {Miyato, Takeru and Kataoka, Toshiki and Koyama, Masanori and Yoshida, Yuichi},
  booktitle = {International Conference on Learning Representations (ICLR)},
  year      = {2018}
}

@inproceedings{ronneberger2015unet,
  title     = {{U-Net}: convolutional networks for biomedical image segmentation},
  author    = {Ronneberger, Olaf and Fischer, Philipp and Brox, Thomas},
  booktitle = {Medical Image Computing and Computer Assisted Intervention -- {MICCAI} 2015},
  series    = {Lecture Notes in Computer Science},
  volume    = {9351},
  pages     = {234--241},
  publisher = {Springer, Cham},
  year      = {2015},
  doi       = {10.1007/978-3-319-24574-4_28}
}

@misc{ulyanov2016instance,
  title  = {Instance normalization: the missing ingredient for fast stylization},
  author = {Ulyanov, Dmitry and Vedaldi, Andrea and Lempitsky, Victor},
  year   = {2016},
  eprint = {1607.08022},
  archivePrefix = {arXiv}
}

@article{wang2004ssim,
  title   = {Image quality assessment: from error visibility to structural similarity},
  author  = {Wang, Zhou and Bovik, Alan C. and Sheikh, Hamid R. and Simoncelli, Eero P.},
  journal = {IEEE Transactions on Image Processing},
  volume  = {13},
  number  = {4},
  pages   = {600--612},
  year    = {2004},
  doi     = {10.1109/TIP.2003.819861}
}

@inproceedings{zhang2018lpips,
  title     = {The unreasonable effectiveness of deep features as a perceptual metric},
  author    = {Zhang, Richard and Isola, Phillip and Efros, Alexei A. and Shechtman, Eli and Wang, Oliver},
  booktitle = {Proceedings of the IEEE Conference on Computer Vision and Pattern Recognition (CVPR)},
  pages     = {586--595},
  year      = {2018},
  doi       = {10.1109/CVPR.2018.00068}
}

@article{gretton2012mmd,
  title   = {A kernel two-sample test},
  author  = {Gretton, Arthur and Borgwardt, Karsten M. and Rasch, Malte J. and Sch{\"o}lkopf, Bernhard and Smola, Alexander},
  journal = {Journal of Machine Learning Research},
  volume  = {13},
  pages   = {723--773},
  year    = {2012}
}

@article{lin1989ccc,
  title   = {A concordance correlation coefficient to evaluate reproducibility},
  author  = {Lin, Lawrence I-Kuei},
  journal = {Biometrics},
  volume  = {45},
  number  = {1},
  pages   = {255--268},
  year    = {1989},
  doi     = {10.2307/2532051}
}

@article{shrout1979icc,
  title   = {Intraclass correlations: uses in assessing rater reliability},
  author  = {Shrout, Patrick E. and Fleiss, Joseph L.},
  journal = {Psychological Bulletin},
  volume  = {86},
  number  = {2},
  pages   = {420--428},
  year    = {1979},
  doi     = {10.1037/0033-2909.86.2.420}
}

@article{menze2015brats,
  title   = {The multimodal brain tumor image segmentation benchmark ({BRATS})},
  author  = {Menze, Bjoern H. and Jakab, Andras and Bauer, Stefan and Kalpathy-Cramer, Jayashree and Farahani, Keyvan and Kirby, Justin and Burren, Yuliya and Porz, Nicole and Slotboom, Johannes and Wiest, Roland and Lanczi, Levente and Gerstner, Elizabeth and Weber, Marc-Andre and Arbel, Tal and Avants, Brian B. and Ayache, Nicholas and Buendia, Patricia and Collins, D. Louis and Cordier, Nicolas and Corso, Jason J. and Criminisi, Antonio and Das, Tilak and Delingette, Herv{\'e} and Demiralp, {\c{C}}a{\u{g}}atay and Durst, Christopher R. and Dojat, Michel and Doyle, Senan and Festa, Joana and Forbes, Florence and Geremia, Ezequiel and Glocker, Ben and Golland, Polina and Guo, Xiaotao and Hamamci, Andac and Iftekharuddin, Khan M. and Jena, Raj and John, Nigel M. and Konukoglu, Ender and Lashkari, Danial and Mariz, Jos{\'e} Antonio and Meier, Raphael and Pereira, S{\'e}rgio and Precup, Doina and Price, Stephen J. and Raviv, Tammy Riklin and Reza, Syed M. S. and Ryan, Michael and Sarikaya, Duygu and Schwartz, Lawrence and Shin, Hoo-Chang and Shotton, Jamie and Silva, Carlos A. and Sousa, Nuno and Subbanna, Nagesh K. and Szekely, Gabor and Taylor, Thomas J. and Thomas, Owen M. and Tustison, Nicholas J. and Unal, Gozde and Vasseur, Flor and Wintermark, Max and Ye, Dong Hye and Zhao, Liang and Zhao, Binsheng and Zikic, Darko and Prastawa, Marcel and Reyes, Mauricio and Van Leemput, Koen},
  journal = {IEEE Transactions on Medical Imaging},
  volume  = {34},
  number  = {10},
  pages   = {1993--2024},
  year    = {2015},
  doi     = {10.1109/TMI.2014.2377694}
}

@article{bakas2017advancing,
  title   = {Advancing {The Cancer Genome Atlas} glioma {MRI} collections with expert segmentation labels and radiomic features},
  author  = {Bakas, Spyridon and Akbari, Hamed and Sotiras, Aristeidis and Bilello, Michel and Rozycki, Martin and Kirby, Justin S. and Freymann, John B. and Farahani, Keyvan and Davatzikos, Christos},
  journal = {Scientific Data},
  volume  = {4},
  pages   = {170117},
  year    = {2017},
  doi     = {10.1038/sdata.2017.117}
}

@article{bakas2022upenn,
  title   = {The {University of Pennsylvania} glioblastoma ({UPenn-GBM}) cohort: advanced {MRI}, clinical, genomics, \& radiomics},
  author  = {Bakas, Spyridon and Sako, Chiharu and Akbari, Hamed and Bilello, Michel and Da, Xiao and Rustam, Sana and McGill, Allison and Kirschbaum, Tom and Rudie, Jeffrey D. and Davatzikos, Christos},
  journal = {Scientific Data},
  volume  = {9},
  number  = {1},
  pages   = {453},
  year    = {2022},
  doi     = {10.1038/s41597-022-01560-7}
}

@article{zwanenburg2020ibsi,
  title   = {The image biomarker standardization initiative: standardized quantitative radiomics for high-throughput image-based phenotyping},
  author  = {Zwanenburg, Alex and Valli{\`e}res, Martin and Abdalah, Mahmoud A. and Aerts, Hugo J. W. L. and Andrearczyk, Vincent and Apte, Aditya and Ashrafinia, Saeed and Bakas, Spyridon and Beukinga, Roelof J. and Boellaard, Ronald and Bogowicz, Marta and Boldrini, Luca and Buvat, Ir{\`e}ne and Cook, Gary J. R. and Davatzikos, Christos and Depeursinge, Adrien and Desseroit, Marie-Charlotte and Dinapoli, Nicola and Dinh, Cu Vinh and Echegaray, Sebastian and El Naqa, Issam and Fedorov, Andrey Y. and Gatta, Roberto and Gillies, Robert J. and Goh, Vicky and G{\"o}tz, Michael and Guckenberger, Matthias and Ha, Sang Mo and Hatt, Mathieu and Isensee, Fabian and Lambin, Philippe and Leger, Stefan and Leijenaar, Ralph T. H. and Lenkowicz, Jacopo and Lippert, Fiona and Losneg{\aa}rd, Are and Maier-Hein, Klaus H. and Morin, Olivier and M{\"u}ller, Henning and Napel, Sandy and Nioche, Christophe and Orlhac, Fanny and Pati, Sarthak and Pfaehler, Elisabeth A. G. and Rahmim, Arman and Rao, Arvind U. K. and Scherer, Jonas and Siddique, Md Minhazul and Sijtsema, Nanna M. and Socarras Fernandez, Jairo and Spezi, Emiliano and Steenbakkers, Roel J. H. M. and Tanadini-Lang, Stephanie and Thorwarth, Daniela and Troost, Esther G. C. and Upadhaya, Taman and Valentini, Vincenzo and van Dijk, Lisanne V. and van Griethuysen, Joost and van Velden, Floris H. P. and Whybra, Philip and Richter, Christoph and L{\"o}ck, Steffen},
  journal = {Radiology},
  volume  = {295},
  number  = {2},
  pages   = {328--338},
  year    = {2020},
  doi     = {10.1148/radiol.2020191145}
}

@article{tustison2010n4itk,
  title   = {{N4ITK}: improved {N3} bias correction},
  author  = {Tustison, Nicholas J. and Avants, Brian B. and Cook, Philip A. and Zheng, Yuanjie and Egan, Alexander and Yushkevich, Paul A. and Gee, James C.},
  journal = {IEEE Transactions on Medical Imaging},
  volume  = {29},
  number  = {6},
  pages   = {1310--1320},
  year    = {2010},
  doi     = {10.1109/TMI.2010.2046908}
}

@article{ben2010theory,
  title   = {A theory of learning from different domains},
  author  = {Ben-David, Shai and Blitzer, John and Crammer, Koby and Kulesza, Alex and Pereira, Fernando and Vaughan, Jennifer Wortman},
  journal = {Machine Learning},
  volume  = {79},
  number  = {1-2},
  pages   = {151--175},
  year    = {2010},
  doi     = {10.1007/s10994-009-5152-4}
}

@inproceedings{kingma2015adam,
  title     = {Adam: {A} Method for Stochastic Optimization},
  author    = {Kingma, Diederik P. and Ba, Jimmy},
  booktitle = {International Conference on Learning Representations ({ICLR})},
  editor    = {Bengio, Yoshua and LeCun, Yann},
  year      = {2015},
  url       = {https://arxiv.org/abs/1412.6980}
}

@article{isensee2019hdbet,
  title     = {Automated brain extraction of multisequence {MRI} using artificial neural networks},
  author    = {Isensee, Fabian and Schell, Marianne and Pflueger, Irada and Brugnara, Gianluca and Bonekamp, David and Neuberger, Ulf and Wick, Antje and Schlemmer, Heinz-Peter and Heiland, Sabine and Wick, Wolfgang and Bendszus, Martin and Maier-Hein, Klaus H. and Kickingereder, Philipp},
  journal   = {Human Brain Mapping},
  volume    = {40},
  number    = {17},
  pages     = {4952--4964},
  year      = {2019},
  doi       = {10.1002/hbm.24750}
}

@article{rohlfing2010sri24,
  title     = {The {SRI24} Multichannel Atlas of Normal Adult Human Brain Structure},
  author    = {Rohlfing, Torsten and Zahr, Natalie M. and Sullivan, Edith V. and Pfefferbaum, Adolf},
  journal   = {Human Brain Mapping},
  volume    = {31},
  number    = {5},
  pages     = {798--819},
  year      = {2010},
  doi       = {10.1002/hbm.20906}
}

@inproceedings{he2016resnet,
  title     = {Deep Residual Learning for Image Recognition},
  author    = {He, Kaiming and Zhang, Xiangyu and Ren, Shaoqing and Sun, Jian},
  booktitle = {Proceedings of the {IEEE} Conference on Computer Vision and Pattern Recognition ({CVPR})},
  pages     = {770--778},
  year      = {2016},
  doi       = {10.1109/CVPR.2016.90}
}

@article{vandermaaten2008tsne,
  title   = {Visualizing Data using {t-SNE}},
  author  = {van der Maaten, Laurens and Hinton, Geoffrey},
  journal = {Journal of Machine Learning Research},
  volume  = {9},
  number  = {86},
  pages   = {2579--2605},
  year    = {2008},
  url     = {https://www.jmlr.org/papers/v9/vandermaaten08a.html}
}

@article{vangriethuysen2017pyradiomics,
  title     = {Computational Radiomics System to Decode the Radiographic Phenotype},
  author    = {van Griethuysen, Joost J. M. and Fedorov, Andriy and Parmar, Chintan and Hosny, Ahmed and Aucoin, Nicole and Narayan, Vivek and Beets-Tan, Regina G. H. and Fillon-Robin, Jean-Christophe and Pieper, Steve and Aerts, Hugo J. W. L.},
  journal   = {Cancer Research},
  volume    = {77},
  number    = {21},
  pages     = {e104--e107},
  year      = {2017},
  doi       = {10.1158/0008-5472.CAN-17-0339}
}

@inproceedings{roth2014twopointfived,
  title     = {A New 2.5{D} Representation for Lymph Node Detection Using Random Sets of Deep Convolutional Neural Network Observations},
  author    = {Roth, Holger R. and Lu, Le and Seff, Ari and Cherry, Kevin M. and Hoffman, Joanne and Wang, Shijun and Liu, Jiamin and Turkbey, Evrim and Summers, Ronald M.},
  booktitle = {Medical Image Computing and Computer-Assisted Intervention -- {MICCAI} 2014},
  series    = {Lecture Notes in Computer Science},
  volume    = {8673},
  pages     = {520--527},
  year      = {2014},
  publisher = {Springer, Cham},
  doi       = {10.1007/978-3-319-10404-1_65}
}

@inproceedings{luo2016erf,
  title     = {Understanding the effective receptive field in deep convolutional neural networks},
  author    = {Luo, Wenjie and Li, Yujia and Urtasun, Raquel and Zemel, Richard},
  booktitle = {Advances in Neural Information Processing Systems (NeurIPS)},
  pages     = {4898--4906},
  year      = {2016}
}

\end{document}